\documentclass[sigconf]{acmart}

\usepackage{graphicx}
\usepackage{amsmath}
\usepackage{amssymb}
\usepackage{amsfonts}
\usepackage{algorithm}
\usepackage{algpseudocode}
\usepackage{multirow}
\usepackage{siunitx}
\usepackage{makecell}
\usepackage{caption}
\usepackage{amsmath}
\usepackage{breqn}
\usepackage{multirow}
\usepackage{makecell}
\usepackage{subcaption}
\usepackage{url}
\usepackage{balance}

\def\BibTeX{{\rm B\kern-.05em{\sc i\kern-.025em b}\kern-.08emT\kern-.1667em\lower.7ex\hbox{E}\kern-.125emX}}
\newcommand{\nn}{\num[group-separator={,}]}

\setcopyright{acmcopyright}

\newcommand{\etal}{\textit{et al}.}

\begin{document}
\copyrightyear{2019}
\acmYear{2019}
\acmConference[MM '19]{Proceedings of the 27th ACM International Conference on Multimedia}{October 21--25, 2019}{Nice, France}
\acmBooktitle{Proceedings of the 27th ACM International Conference on Multimedia (MM '19), October 21--25, 2019, Nice, France}
\acmPrice{15.00}
\acmDOI{10.1145/3343031.3350887}
\acmISBN{978-1-4503-6889-6/19/10}

\fancyhead{}

\title{MetaAdvDet: Towards Robust Detection of Evolving Adversarial Attacks}

\author{Chen Ma}
\email{sharpstill@163.com}
\orcid{0000-0001-6876-3117}
\affiliation{%
	\institution{School of Software, Tsinghua University \& Beijing National Research Center for Information Science and Technology (BNRist)}
	\city{Beijing}
	\country{China}
	\postcode{100084}
}

\author{Chenxu Zhao}
\email{zhaochenxu1@jd.com}
\affiliation{%
	\institution{JD AI Research}
	\city{Beijing}
	\country{China}
	\postcode{100101}
}

\author{Hailin Shi}
\email{shihailin@jd.com}
\affiliation{%
	\institution{JD AI Research}
	\city{Beijing}
	\country{China}
	\postcode{100101}
}

\author{Li Chen}
\email{chenlee@tsinghua.edu.cn}
\authornote{Corresponding author: Li Chen}
\affiliation{%
	\institution{School of Software, Tsinghua University \& BNRist}
	\city{Beijing}
	\country{China}
	\postcode{100084}
}

\author{Junhai Yong}
\email{yongjh@tsinghua.edu.cn}
\affiliation{%
	\institution{School of Software, Tsinghua University \& BNRist}
	\city{Beijing}
	\country{China}
	\postcode{100084}
}

\author{Dan Zeng}
\email{dzeng@shu.edu.cn}
\affiliation{
	\institution{Shanghai University}
	\city{Shanghai}
	\country{China}
}

%
\renewcommand{\shortauthors}{Ma, et al.}

\begin{abstract}
Deep neural networks (DNNs) are vulnerable to adversarial attack which is maliciously implemented by adding human-imperceptible perturbation to images and thus leads to incorrect prediction. 
Existing studies have proposed various methods to detect the new adversarial attacks.
However, new attack methods keep evolving constantly and yield new adversarial examples to bypass the existing detectors. It needs to collect tens of thousands samples to train detectors, while the new attacks evolve much more frequently than the high-cost data collection. Thus, this situation leads the newly evolved attack samples to remain in small scales.
To solve such few-shot problem with the evolving attacks, we propose a meta-learning based robust detection method to detect new adversarial attacks with limited examples. 
Specifically, the learning consists of a double-network framework:  a task-dedicated network and a master network which alternatively learn the detection capability for either seen attack or a new attack.
To validate the effectiveness of our approach, we construct the benchmarks with few-shot-fashion protocols based on three conventional datasets, i.e. CIFAR-10, MNIST and Fashion-MNIST. Comprehensive experiments are conducted on them to verify the superiority of our approach with respect to the traditional adversarial attack detection methods. The implementation code is available online at \url{https://github.com/sharpstill/MetaAdvDet}.
\end{abstract}

%
%

\ccsdesc[500]{Computing methodologies~Computer vision problems}

%
\keywords{adversarial example detection, meta-learning, few-shot learning, evolving adversarial attacks}

%

%

\maketitle
\begin{figure}[htbp]
	\setlength{\abovecaptionskip}{0pt}%
	\setlength{\belowcaptionskip}{0pt}%
	\begin{center}
		\includegraphics[width=1\linewidth]{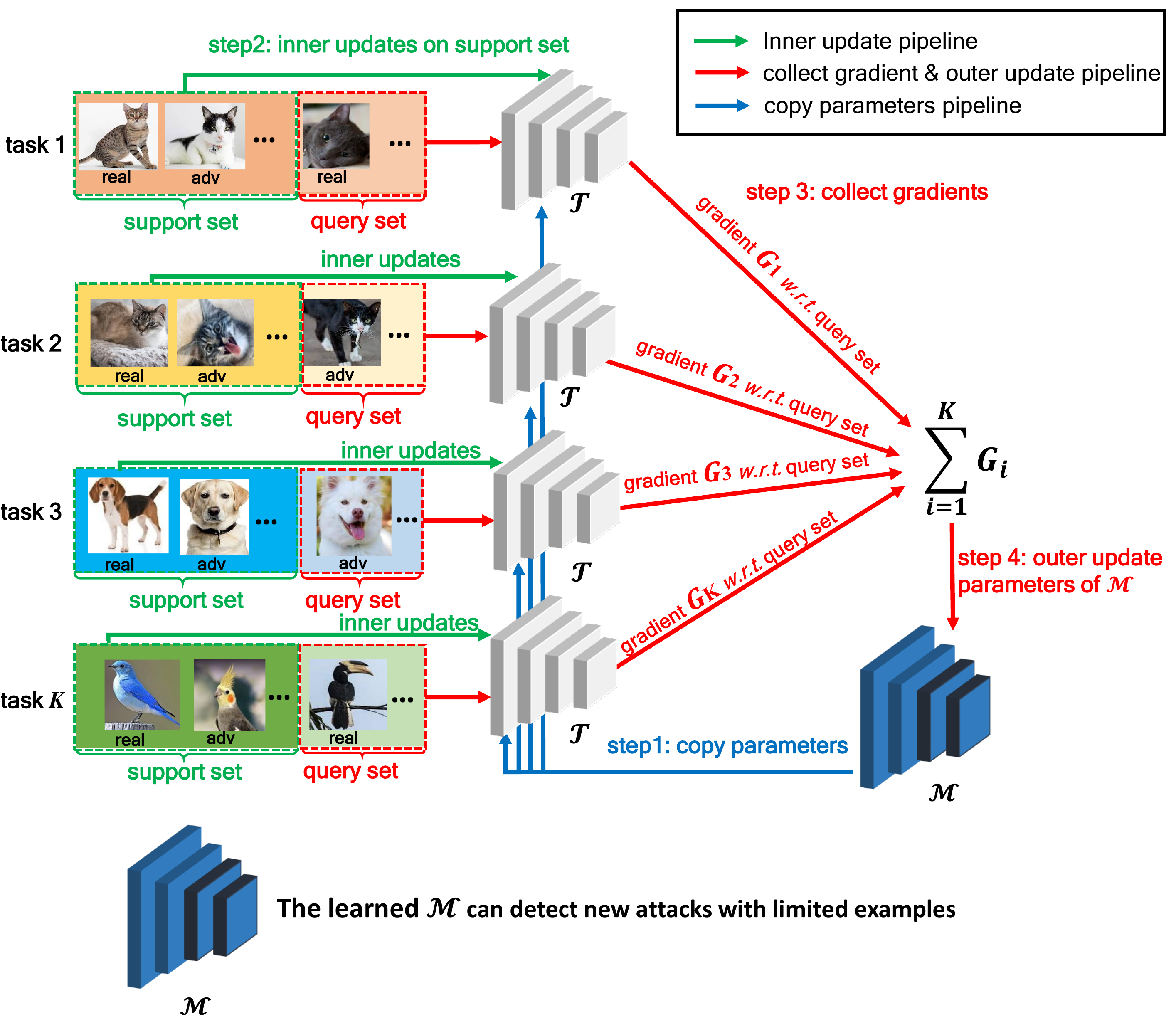}
	\end{center}
	\caption{The procedure of MetaAdvDet training in one mini-batch (best viewed in color). The approach consists of a double-network framework: $\mathcal{M}$ and $\mathcal{T}$. $\mathcal{T}$ is the task-dedicated network which focuses on learning each task. It copies parameters from master network $\mathcal{M}$ at the beginning, and then trains on the support set. After a couple of iterations (inner update step), $\mathcal{T}$ converges and computes the gradient $G_i$ on the query set of task $i$. $\mathcal{M}$ accumulates the gradients $\sum_{i=1}^{K} G_{i}$ to update its parameters $\mathcal{M}_\theta$ which is prepared for the next mini-batch learning. The learned $\mathcal{M}$ can be used to detect new attacks with limited new samples.  More details can be found in Sec.\ref{sec:approach} and Algorithm \ref{alg:trn}.}
	
	\label{fig:fig1}
	\vspace{-0.4cm}
\end{figure}
\section{Introduction}

The evolving adversarial attacks threaten the deep convolutional neural networks (DNNs) via adding human-imperceptible perturbation to clean images and thus lead to incorrect prediction. Various defense methods have been proposed for detecting attacks, which distinguish adversarial images and real images via capturing the features of DNNs under attacks \cite{ma2018characterizing,bhagoji2017dimensionality,tian2018detecting,xu2017feature}. However, new attack methods keep constantly evolving and yield new adversarial examples to bypass existing detector. For example, C\&W attack \cite{Carlini2017TowardsET} is proposed to circumvent all existing detection techniques at that time. Certain detection techniques have been proposed to detect new attacks \cite{sorin2002safetynet,dathathri2018detecting}, these techniques are promising. However, most of them need tens of thousands of examples to train which are infeasible in practice. Because new attacks evolve much faster than the high-cost data collection, which results in a few-shot learning problem with evolving attacks. This issue makes the detection of adversarial examples still challenging.

Therefore, we study on how to tackle such few-shot learning problem, and propose a meta-learning based training approach with the learning-to-learn strategy. It focuses on learning to detect new attack from one or few instances of that attack. We name our approach as MetaAdvDet, refers to \textbf{Meta}-learning \textbf{Adv}ersarial \textbf{Det}ection approach. To this end, the approach is equipped with a double-network framework for learning from \textit{tasks}, which is defined as the small data collection with real examples and randomly chosen type of attacks. The purpose of introducing the tasks is to simulate new attack scenarios. To better learn from tasks, MetaAdvDet uses one network to focus on learning individual tasks, and the other network to learn the general detection strategy over multiple tasks. Fig. \ref{fig:fig1} illustrates the training procedure of one mini-batch, more details are described in Sec. \ref{sec:approach}. Each task is divided into \textit{support set} and \textit{query set}, which are used for learning either basic detection capability on old attacks, or minimizing the \textit{test} error on new attacks. After training, the framework efficiently detects new attack with fine-tuning on limited examples. In contrast, the DNN based methods that use` the traditional training approach perform much worse in detecting new attacks than ours.

To comprehensively validate the detection techniques in terms of evolving attacks, we propose evaluations in following dimensions to validate the superiority of our approach in the few-shot problem.

\textbf{Cross-adversary Dimension}. To assess the capability of detecting new types of attacks in test set with few-shot samples..

\textbf{Cross-domain Dimension}. To assess the capability of detecting all attacks across different domains with few-shot samples.

\textbf{Cross-architecture Dimension}. To assess the capability of detecting the adversarial examples that are generated by attacking the classifier with new architecture.

\textbf{White-box attack dimension}. To assess the capability of detecting white-box attacks with few-shot samples.

To validate the effectiveness of our approach from above dimensions, we propose benchmarks with the few-shot-fashion protocol on three conventional datasets, \textit{i.e.} CIFAR-10, MNIST and Fashion-MNIST datasets. The benchmarks include the generated adversarial examples by using various types of attacks, and it also defines the partition of train set and test set to simulate the scenario of testing the evolving attack's detection.

In experiments, we compare our approach with end-to-end state-of-the-art methods using these benchmarks, and the results show that our approach surpasses the existing method by a large margin.

We summarize the main contributions below:

(1) To the best of our knowledge, we are the first to define the adversarial attack detection problem as a few-shot learning problem of detecting evolving new attacks.

(2) We propose a meta-learning based approach: MetaAdvDet, it is equipped with a double-network framework with the learning-to-learn strategy for detecting evolving attacks. Benefiting from the learning-to-learn strategy, our approach is able to achieve high performance in detecting new attacks. 

(3) To comprehensively validate our approach in terms of evolving attacks, we construct benchmarks with the few-shot-fashion protocol on three datasets, \textit{i.e.} CIFAR-10, MNIST and Fashion-MNIST. The benchmarks define the partition of train set and test set to simulate the scenario of testing the evolving attack. We believe the proposed benchmark is useful for the future research of defending evolving attacks.

\section{Background}
Many attempts have been made to detect or defense against adversarial attack. We first introduce the defense techniques, and then we introduce the meta-learning techniques that related to our work.

\vspace{-4pt}
\subsection{Defense Techniques}
The \textit{adversary} algorithm is used to generate the adversarial examples which makes the classifier to output incorrect prediction. 
Many defense techniques have been proposed to defend against adversarial attack, these techniques generally fall into two categories. 

The first category attempts to build a robust model that classifies the adversarial example correctly, such as \cite{papernot2016distillation,akhtar2018defense,song2018pixeldefend,liao2018defense}. However, certain new attacks \cite{chen2017zoo,li2019nattack} are deliberately implemented to grasp the weakness of these methods to circumvent the defense. For example, Athalye \textit{et al.} \cite{obfuscated-gradients} identifies the obfuscated gradients, which is a kind of gradient masking, that leads to a false sense of security in defenses. Based on their findings, the new attacks are proposed to circumvent 7 of 9 defenses relying on obfuscated gradients.

Due to the difficulty, the second category of defense techniques turn to distinguish the adversarial examples from real ones, in order to improve security and detect malicious users. This category refers to adversarial attack detection. Unlike the first category, adversarial detection does not need to classify the adversarial image correctly, but only to identify them. Essentially, a detector is also a binary classifier which is trained on the real and adversarial examples. Based on this idea,
certain detection techniques \cite{carrara2017detecting,sorin2002safetynet, metzen2017detecting} build a subnet classifier to capture the hidden layer's features of the adversarial example. Other methods include (1) capturing the difference of DNN's output between real and adversarial images when applying certain transformation to the input images \cite{tian2018detecting,xu2017feature,dathathri2018detecting,bhagoji2017dimensionality}, (2) utilizing the intrinsic dimensionality of adversarial regions \cite{ma2018characterizing}, (3) employing new loss function to encourage DNN to learn a more distinguishable representation \cite{pang2018towards,wan2018rethinking}, (4) using statistical test \cite{grosse2017statistical}, and (5) using the capsule network \cite{frosst2018darccc}.

However, the high-cost data collection cannot keep up with the evolution frequency of the attacks, which leads the training for detecting new attacks difficulty. For example, when a new attack first appears without publishing the source code, most of defenders have insufficient examples to train the detector. This situation makes the issue of detecting evolving attacks highly urgent. We categorize this issue as a new defense problem, which is a few-shot learning problem of detecting evolving attacks.

\subsection{Meta-Learning} 
Few-shot learning problem \cite{vinyals2016matching,snell2017prototypical} has been studied for a long time, which is defined as learning from few samples. The meta-learning techniques \cite{finn2017model,li2017meta,RecastingMAML,mishra2018a,Jamal_2019_CVPR} are promising for addressing the few-shot learning problem, which usually trains a meta-learner on the distribution of few-shot tasks so that the it can generalize and perform well on the unseen task. Model-agnostic meta-learning (MAML) \cite{finn2017model} is a typical meta-learning approach, which learns a internal representation that is widely suitable for many tasks. It learns a proper weight initialization on the support set and then updates itself to perform well on the query set. To update the weights more efficiently, Meta-SGD \cite{li2017meta} makes the meta-learner not only to learn the weight initialization but also update direction and learning rate. For better understanding in this field, we introduce the terminologies of meta-learning, as describe below.

\textbf{Task}: A meta-learning model (meta-learner) should be trained over a variety of tasks and optimized for the best performance on the task distribution, including potentially unseen
tasks. The concept of ``task'' in this paper is totally different from the concept of ``multi-task learning'', but only a manner of data partition that the meta-learner used to train. 

\textbf{Support\&query set}: Each task is split into two subsets, which are the support set for learning the basic classification on old tasks, and the query set for training in the train stage or testing in the test stage. It should be emphasized that the support set and query set from the same task have the same data distribution.

\textbf{Way} is the class in each task that the meta-learner wish to discriminate, whose number may be specified arbitrarily and do not need to equal the ground truth class number. 

\textbf{Shot} is the number of samples in each way of the support set. For example, an $N$-way, $K$-shot classification task includes the support set with $K$ labeled examples for each of $N$ classes. 

Based on the spirit of meta-learning, we propose the training method with a double-network framework and introduce the double-update scheme for achieving fast adaption capacity. Experiments show the superiority of our approach in detecting new attacks.

\section{Approach}

\subsection{Overview}

\begin{figure}[tb]
	\setlength{\abovecaptionskip}{0pt}%
	\setlength{\belowcaptionskip}{0pt}%
	\begin{center}
		\includegraphics[width=1\linewidth]{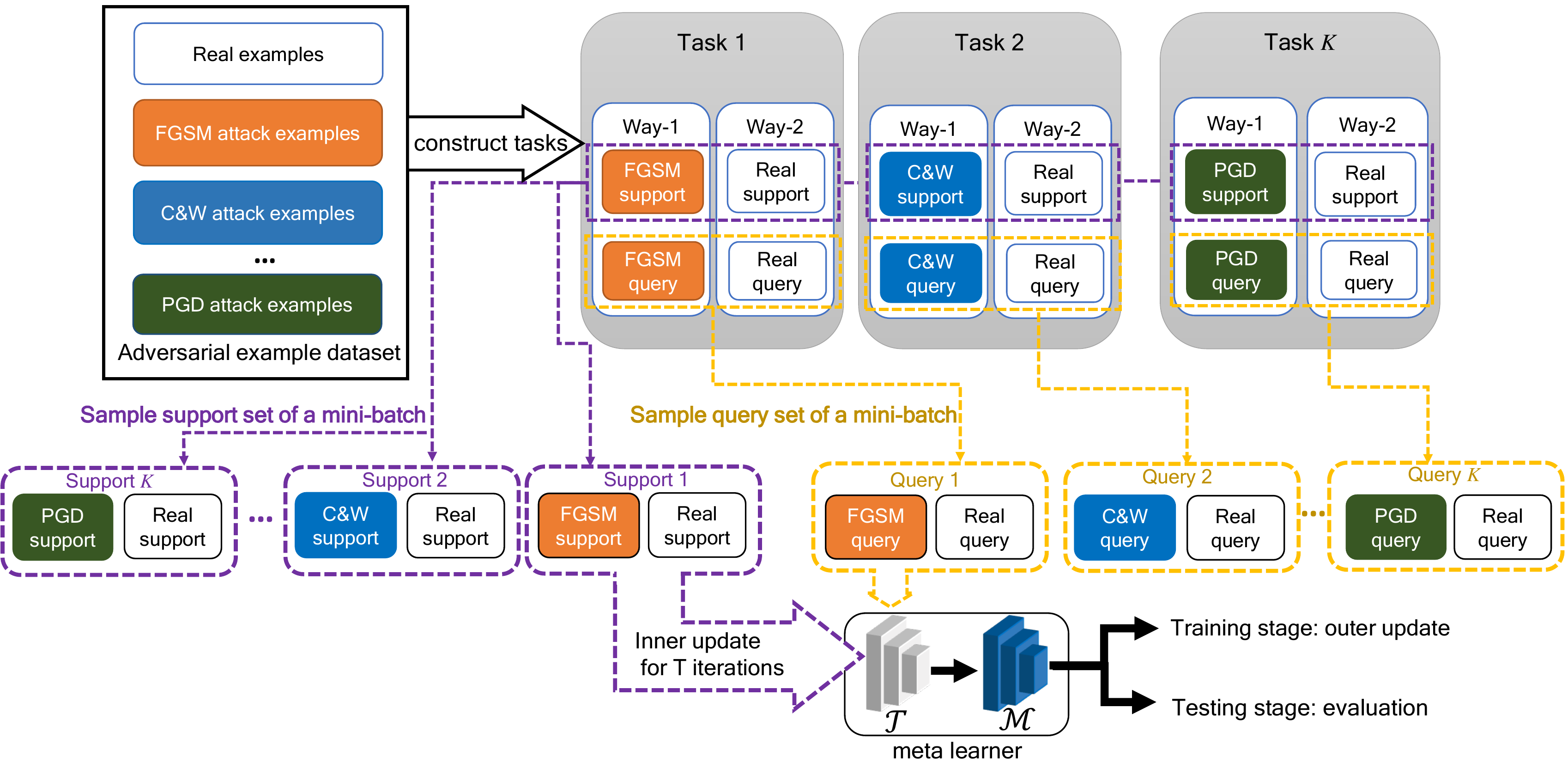}
	\end{center}
	\caption{The details of constructing tasks for training and testing the meta-learner (including $\mathcal{T}$ and $\mathcal{M}$). Each task (support\&query set) is sampled independently from the dataset, and each mini-batch consists of $K$ tasks. The training employs a double-update scheme: inner update and outer update. The inner update represents that $\mathcal{T}$ learns on support set, and the outer update represents $\mathcal{M}$ updates with the accumulated gradients of $\mathcal{T}$ on query set (as described in Fig. \ref{fig:fig1} and Sec. \ref{sec:approach}). $\mathcal{M}$ can be used to detect new attacks.}
	\label{fig:arch}
	\vspace{-0.4cm}
\end{figure}

The evolving adversarial attacks are hard to distinguish due to the insufficient new adversarial examples for training the detector, results in the few-shot learning problem. One of the keys for solving this problem is to use the power of meta-learning techniques. 
Typical meta-learning methods (\textit{e.g.} MAML \cite{finn2017model}) are trained for learning the task distribution. Because the categories of data in each task are randomly chosen, the meta-leaner acquires fast adaption capability to unseen data type via learning these tasks. To model the attack detection technique in the meta-learning style framework, we collect various types of attacks to construct adversarial example dataset into the multiple tasks form (Fig. \ref{fig:arch}). Each task is a small data collection with a randomly chosen attack which represents one attacking scenario, so the large amount of tasks make the meta-learner experience various attacking scenarios, so that it can adapt to new attacks rapidly. Our approach is equipped with a double-network framework with learning-to-learn strategy, which focuses on learning how to learn new tasks faster by reusing previous experience, rather than considering new tasks in isolation. Specifically, one network of our framework focuses on learning from individual tasks (named task-dedicated network $\mathcal{T}$), the other network updates its parameters based on the gradient accumulated from the $\mathcal{T}$ (named master network $\mathcal{M}$), to learn a general strategy over all tasks (Fig. \ref{fig:fig1}). This double-network framework leads to the double update scheme, corresponding to the two networks. 

Fig. \ref{fig:arch} shows the details of constructing tasks for training and testing the meta-learner, Fig. \ref{fig:fig1} demonstrates the procedure of training in one mini-batch, detailed steps are shown in Algorithm \ref{alg:trn}.

\subsection{Learning MetaAdvDet}
\label{sec:approach}
As we mentioned earlier, the learning-to-learn strategy is proposed to learn new attacks by reusing previous experience of detecting old attacks. Following the typical setting of meta-learning, all the training data are organized into tasks, each task is divided to two subsets, namely, the support set for learning basic capability of detecting old attacks, and the query set acts as the surrogate of new attacks for achieving rapid adaption in detecting new attacks of test set.   
To learn the tasks, the meta-learner includes a double-network framework, \textit{i.e.} the master network $\mathcal{M}$, and a task-dedicated network $\mathcal{T}$ which is cloned from $\mathcal{M}$ to learn from individual tasks. $\mathcal{T}$ updates its parameters $\mathcal{T}_\theta$ based on each task's support set, and then it calculates its gradient of the query set, which will be accumulated to update $\mathcal{M}$'s parameters $\mathcal{M}_\theta$ (Fig. \ref{fig:fig1}). The same $\mathcal{M}_\theta$ will be copied and overwritten to the $\mathcal{T}_\theta$ before learning next task. The $\mathcal{M}$ and $\mathcal{T}$ output the classification probability to distinguish the real and adversarial example, corresponding to the two-way configuration. The two-way configuration stipulates that one of the ways should use real examples in all tasks. Two options are considered, \textit{i.e.} the randomized-way setting, whose two-way labels are shuffled in each task; and the fixed-way setting, which uses label 1 for real example and label 0 for adversarial example in all tasks. We will compare the effect of above two options in Sec. \ref{sec:ablation_study}.

Algorithm \ref{alg:trn} shows the training procedure, Fig. \ref{fig:fig1} shows the detail of one mini-batch training procedure. $\mathcal{T}$ copies the all the parameters from $\mathcal{M}$ at the beginning of learning task $\mathbb{T}_i$, where the subscript $i$ denotes the task index. Then, the inner update step updates the parameters of $\mathcal{T}$ by using the support set of $\mathbb{T}_i$ for multiple iterations. Line \ref{line:forward_support} and line \ref{line:inner_update} demonstrate this step, which is the same with the supervised learning in the traditional DNN: we directly feed input images to $\mathcal{T}$ and uses the gradient descent to update its parameters based on the classification ground truth. Unlike existing methods that applying transformation on input images \cite{tian2018detecting,xu2017feature}, we should note that the input image is not applied any transformation in this step of our approach. Finally, the meta-learner acquires rapid adaption capability by considering to minimize the \textit{test error} on new data, this is the role that the outer update step upon the query set plays. More specifically, we calculate the cross entropy loss $\mathcal{L}$ on the query set of task $\mathbb{T}_i$ to obtain the gradient $G_i$ \textit{w.r.t.} $\mathcal{T}_\theta$, which is accumulated from learning all tasks $\mathbb{T}_1,\dotsc,\mathbb{T}_K$ and finally sent to $\mathcal{M}$. Because $\mathcal{M}$ and $\mathcal{T}$ use the same network structure and parameters, the accumulated gradient can be used to update parameters of $\mathcal{M}$, namely $\mathcal{M}_\theta$. Thus, $\sum_{i=1}^K G_i$ updates the $\mathcal{M}_\theta$ for learning the strategy over the multi-task distribution. 

\begin{algorithm}
	\caption{\small MetaAdvDet training procedure}
	\label{alg:trn} 
	\begin{algorithmic}[tb] 
		\Require
		master network $\mathcal{M}$ and its parameters $\mathcal{M}_\theta$, task-dedicated network $\mathcal{T}$ and its parameters $\mathcal{T}_\theta$, the feed-forward function $f_{\mathcal{T}_{\theta}}$ of $\mathcal{T}$, max iterations $N$, inner-update learning rate $\lambda_1$, outer-update learning rate $\lambda_2$, inner updates iteration $T$, the multi-task format dataset $\mathcal{D}$, cross entropy loss function $\mathcal{L}$.
		\Ensure the learned network $\mathcal{M}$
	\end{algorithmic}
	\begin{algorithmic}[1] 
		\For{$iter \gets 1$ to $N$}
		\State sample $K$ tasks $\mathbb{T}_{i, i\in\{1,\cdots, K\}}$ from $\mathcal{D}$
		\For{$i \gets 1$ to $K$}
		\State $S_i$ and $Q_i$ $\gets$ support set and query set of $\mathbb{T}_i$
		\State $\mathcal{T}_\theta \gets \mathcal{M}_\theta$\Comment{copy parameters from $\mathcal{M}$ to $\mathcal{T}$}
		\State $\mathcal{T}_{\theta^\prime} \gets \mathcal{T}_\theta$\Comment{$\mathcal{T}_\theta$ will be used in the outer update}
		\For{$t \gets 1$ to $T$}
		\State Calculate $\nabla_{\mathcal{T}_{\theta^\prime}} \mathcal{L}(f_{\mathcal{T}_{\theta^\prime}})$ by using $S_i$ \label{line:forward_support}
		\State $\mathcal{T}_{\theta^\prime} \gets \mathcal{T}_{\theta^\prime} - \lambda_1 \nabla_{\mathcal{T}_{\theta^\prime}} \mathcal{L}(f_{\mathcal{T}_{\theta^\prime}})$  \Comment{inner update} \label{line:inner_update}
		\EndFor
		\State $G_i \gets \nabla_{\mathcal{T}_\theta} \mathcal{L}(f_{\mathcal{T}_{\theta^\prime}})$ by using $Q_i$ \label{line:forward_query}
		\EndFor
		\State $\mathcal{M}_\theta \gets \mathcal{M}_\theta - \lambda_2 \sum_{i=1}^K G_i$ \Comment{outer update}
		\EndFor
		\State\Return $\mathcal{M}$
	\end{algorithmic}
\end{algorithm}

Following popular few-shot-fashion testing procedure \cite{miniimagenet}, the evaluation restricts that the method needs to be evaluated on all test tasks. Algorithm \ref{alg:test} shows the testing procedure. The few-shot-fashion testing procedure should include a fine-tune step by using few-shot examples, as shown in line \ref{line:finetune} of Algorithm \ref{alg:test}. In experiments, we adopt a general binary classifier with DNN as the baseline for comparison. DNN uses a single network for training, whereas MetaAdvDet uses a double-network framework to obtain the learning-to-learn strategy. The experiment proves the superiority of our approach in detecting new attacks (Sec. \ref{sec:expr_cross_adv}).  
\begin{algorithm}
	\caption{\small MetaAdvDet testing procedure}
	\label{alg:test} 
	\begin{algorithmic}[tb] 
		\Require
		master network $\mathcal{M}$ and its learned parameters $\mathcal{M}_\theta$, task-dedicated network $\mathcal{T}$ and its parameters $\mathcal{T}_\theta$, the feed-forward function $f_{\mathcal{T}_{\theta}}$ of $\mathcal{T}$, fine-tune iterations $T$, learning rate $\lambda$, test tasks $\mathbb{T}_{i, i\in\{i,\cdots, N\}}$ which is obtained by reorganizing the test set, cross entropy loss $\mathcal{L}$, ground truth $Y_{i, i\in\{i,\cdots,N\}}$ of the query set.
		\Ensure the average F1 score over all tasks
	\end{algorithmic}
	
	\begin{algorithmic}[1] 
		\For{$\mathbb{T}_i \gets \mathbb{T}_1$ to $\mathbb{T}_N$}  \Comment{iterate over all test tasks}
		\State $S_i$ and $Q_i$ $\gets$ support set and the query set of $\mathbb{T}_i$
		\State $\mathcal{T}_\theta \gets \mathcal{M}_\theta$\Comment{copy parameters to ensure each task is tested independently}
		\For{$t \gets 1$ to $T$}
		\State Calculate $\nabla_{\mathcal{T}_\theta} \mathcal{L}(f_{\mathcal{T}_{\theta}})$ by using $S_i$
		\State $\mathcal{T}_{\theta} \gets \mathcal{T}_\theta - \lambda \nabla_{\mathcal{T}_\theta} \mathcal{L}(f_{\mathcal{T}_{\theta}})$  \Comment{fine-tune step}\label{line:finetune}
		\EndFor
		\State $\hat{Y}_i \gets f_{\mathcal{T}_{\theta}}(Q_i)$ \Comment{get prediction of query set of task $i$}
		\State $\text{score}_i \gets \text{F1}(\hat{Y}_i,  Y_i)$
		\EndFor
		\State $\text{F1 score} \gets \frac{1}{N}\sum_{i=1}^{N} score_i$\label{line:avg_sum}
		\State\Return F1 score
	\end{algorithmic}
\end{algorithm}

The F1 score of the query set is adopted as the metric for evaluating the performance of detection techniques (Sec. \ref{sec:eval_metric}). Note that the F1 score is calculated upon individual tasks, the final F1 score is obtained via averaging F1 scores of all tasks, which follows the few-shot-fashion testing procedure of MiniImagenet \cite{miniimagenet}, steps are shown in line \ref{line:avg_sum} of Algorithm \ref{alg:test}. All the compared methods should use this metric and include the fine-tune step for fair comparison.

\section{Proposed Benchmark}
\subsection{Adversarial Example Datasets Construction}
In order to validate the effectiveness of our approach, we construct the adversarial example datasets based on the conventional datasets. The built datasets use fifteen adversaries to yield examples whose data sources come from CIFAR-10 \cite{CIFAR10}, MNIST \cite{MNIST} and Fashion-MNIST \cite{FashionMNIST} datasets, named AdvCIFAR, AdvMNIST and AdvFashionMNIST respectively. To train the detectors for distinguishing real examples and adversarial examples, each dataset includes an additional real example's category whose data are directly transfered from original dataset (\textit{i.e.} CIFAR-10 \textit{etc.}). All fifteen types of adversarial examples are generated by utilizing CleverHans library \cite{papernot2018cleverhans}, which attacks the classifiers with three architectures for each adversary, namely 4 conv-layers network (conv-4), ResNet-10 \cite{he2016deep} and ResNet-18 \cite{he2016deep}. Note that MI-FGSM, BIM and PGD attacks adopt the $\ell_{\infty}$ norm version, C\&W and Deepfool attacks adopt the $\ell_{2}$ norm version. Such adoptions are based on the attack successful rate. In addition, the adversarial examples of L-BFGS attack \cite{szegedy2014intriguing} is used as the validation set. The BPDA attack \cite{obfuscated-gradients} that utilizes the obfuscate gradients of defense is not used, because our approach does not reply on obfuscate gradients. The statistical data for the adversarial example datasets are shown in Tab. \ref{tab:dataset_count}. 

\newcommand{\tabincell}[2]{\begin{tabular}{@{}#1@{}}#2\end{tabular}}
\begin{table}[htp]
	\scriptsize	
	\centering
	\tabcolsep=0.15cm
	\caption{Our adversarial example datasets contain the examples generated by attacking different architectures, including a 4 conv-layers network (conv-4), ResNet-10 and ResNet-18.This table lists the amount of adversarial examples which are generated by successfully attacking the conv-4 network.}
	\begin{tabular}{p{2.6cm}|l|l|l|l|l|l}
		\toprule
		\multirow{2}{2.2cm}{adversary} & \multicolumn{2}{c|}{AdvCIFAR} & \multicolumn{2}{c|}{AdvMNIST} & \multicolumn{2}{c}{AdvFashionMNIST} \\
		\cmidrule(r){2-7}
		&  train & test & train & test & train & test \\
		\midrule
		FGSM \cite{goodfellow6572explaining} & \nn{46851} & \nn{9260} & \nn{23646}& \nn{3853}  & \nn{48368} & \nn{7999}  \\
		MI-FGSM \cite{dong2018boosting}  & \nn{46742} & \nn{9205} & \nn{58445} & \nn{9701} & \nn{56744} & \nn{9362}  \\
		BIM \cite{kurakin2016adversarial} & \nn{46076} & \nn{9118} & \nn{58522} & \nn{9715}  &  \nn{56587} & \nn{9341}   \\
		PGD \cite{madry2018towards} & \nn{47911} & \nn{9504} & \nn{58439} & \nn{9693}  & \nn{57060} & \nn{9428} \\
		C\&W \cite{Carlini2017TowardsET} & \nn{47810} & \nn{9509} & \nn{58121} & \nn{9651}  & \nn{57072} & \nn{9435} \\
		jsma \cite{papernot2016limitations}& \nn{49141} & \nn{9807} & \nn{26377} & \nn{4305} & \nn{39804} & \nn{6770}  \\
		EAD \cite{chen2018ead} & \nn{45146} & \nn{8908} & \nn{59458} & \nn{9885}  & \nn{56157} & \nn{9283} \\
		SPSA \cite{uesato2018adversarial}& \nn{42183} & \nn{8436} & \nn{1260} & \nn{245} & \nn{12604} & \nn{2225} \\
		Spatial Transformation \cite{xiao2018spatially} & \nn{47075} & \nn{9320} & \nn{58820} & \nn{9770} & \nn{59520} & \nn{9917}  \\
		VAT \cite{miyato2016distributional} & \nn{28758} & \nn{5788} & \nn{11392} & \nn{1869} & \nn{24774} & \nn{4159}\\
		semantic \cite{hosseini2017limitation} & \nn{31704} & \nn{6415} & \nn{52398} & \nn{8723}  & \nn{47401} & \nn{7918} \\
		MaxConfidence \cite{goodfellow2019evaluation} & \nn{48293} & \nn{9565} & \nn{57676} & \nn{9604}  & \nn{57309} & \nn{9469}  \\
		Deepfool \cite{moosavi2016deepfool}& \nn{44740} & \nn{8879} & \nn{59461} & \nn{9886}  & \nn{56171} & \nn{9294}  \\
		NewtonFool \cite{jang2017objective} & \nn{45240} & \nn{8916} & \nn{59473} & \nn{9884}  & \nn{56249} & \nn{9294} \\
		\midrule
		L-BFGS \cite{szegedy2014intriguing}(validation set) & \multicolumn{2}{l|}{\nn{9683}} & \multicolumn{2}{l|}{\nn{9583}} & \multicolumn{2}{l}{\nn{9598}} \\
		\bottomrule
	\end{tabular}
	
	\label{tab:dataset_count}
	
\end{table}

\subsection{Cross-Adversary Benchmark}

\label{sec:cross_adv}
To validate the effectiveness of detection techniques in detecting new attacks, we configure the train set and test set contain no common type of adversarial examples to simulate this situation.
To this end, the attacks are grouped based on their categories, and we propose the cross-adversary benchmark which assigns the different adversary groups to the train set and test set.  

\begin{table}[!htb]
	\scriptsize
	\tabcolsep=0.1cm
	\setlength{\abovecaptionskip}{0pt}%
	\setlength{\belowcaptionskip}{0pt}%
	\caption{The definition of adversary groups in the cross-adversary benchmark.}
	\begin{center}
		\begin{tabular}{p{3cm}|p{2.77cm}|p{0.8cm}|p{1.13cm}}
			\toprule
			Train Adversary Group & Test Adversary Group & Validation & Train\&Test \\
			\midrule
			FGSM, MI-FGSM, BIM, PGD, C\&W, jsma, SPSA, VAT, MaxConfidence & EAD, semantic, Deepfool, Spatial Transformation, Newtonfool & L-BFGS & same domain \\
			\bottomrule
		\end{tabular}
	\end{center}
	\vspace{-0.2cm}
	\label{tab:adversary_group}
\end{table}

Tab. \ref{tab:adversary_group} shows the adversary groups of cross-adversary benchmark. The grouping principles of this benchmark are: 
(1) each adversary should be assigned to one group only. (2) The similar adversaries should be assigned into the same group.
For example, the MI-FGSM adversary is a modification of FGSM, and thus they are similar and we make them into one group. Based on this benchmark, the train set and test set should not include attacks of the same group simultaneously. Note that in this benchmark, the adversaries of train group extract the train set of the adversarial example dataset (\textit{e.g.} train set of FGSM in Tab. \ref{tab:dataset_count}) to train the detectors. Similarly, the detectors are evaluated on the test set of the test group's adversaries.	

\subsection{Cross-Domain Benchmark}
\label{sec:cross_domain}
\begin{table}[htb]
	\scriptsize
	\tabcolsep=0.1cm
	\setlength{\abovecaptionskip}{0pt}%
	\setlength{\belowcaptionskip}{0pt}%
	\caption{The cross-domain benchmark consists of 2 protocols on the AdvMNIST and AdvFashionMNIST.}
	\begin{center}
		\begin{tabular}{c|c|c|c|c}
			\toprule
			Protocol & Train Domain & Test Domain & Attack Types & Test Shots \\
			\midrule
			\#1 & AdvMNIST & AdvFashionMNIST & all attacks & 1-shot, 5-shot \\
			\#2 & AdvFashionMNIST & AdvMNIST & all attacks & 1-shot, 5-shot \\
			\bottomrule
		\end{tabular}
	\end{center}
	\vspace{-0.4cm}
	\label{tab:cross_domain_protocol}
\end{table}
The concept of a domain indicates an adversarial dataset, \textit{e.g.} AdvMNIST. Since different domains have different data distributions, which leads the cross-domain benchmark to a more challenging benchmark. To evaluate the capability of detecting the adversarial examples generated from new domain, the detectors are trained one domain (\textit{Train Domain}), and tested on the other domain (\textit{Test Domain}). In this benchmark, we focus on the transferability between two datasets, namely AdvMNIST and AdvFashionMNIST, as listed in Tab. \ref{tab:cross_domain_protocol}. 
Note that in this benchmark, all types of the attacks are used to train the detector.

\subsection{Cross-Architecture Benchmark}
\label{sec:cross_arch}
\begin{table}[!htb]
	\scriptsize
	\tabcolsep=0.1cm
	\setlength{\abovecaptionskip}{0pt}%
	\setlength{\belowcaptionskip}{0pt}%
	\caption{The cross-architecture benchmark consists of 4 protocols, this benchmark indicates the examples of train set and test set are generated by attacking different networks.}
	\begin{center}
		\begin{tabular}{c|c|c|c|c|c}
			\toprule
			Protocol & Train Arch & Test Arch & Attack Types & Test Shots & Train\&Test \\
			\midrule
			\#1 & ResNet-10 & ResNet-18 & all attacks & 1-shot, 5-shot & same domain \\
			\#2 & ResNet-18 & ResNet-10 & all attacks & 1-shot, 5-shot & same domain \\
			\#3 & conv-4 & ResNet-10 & all attacks & 1-shot, 5-shot & same domain\\
			\#4 & ResNet-10 & conv-4 & all attacks & 1-shot, 5-shot & same domain\\
			\bottomrule
		\end{tabular}
	\end{center}
	\vspace{-0.2cm}
	\label{tab:cross_arch_protocol}
\end{table}
Existing studies show that the adversarial examples generated by attacking one architecture can fool another architecture \cite{papernot2016transferability,liu2017delving}. To validate the detection capability in this situation, this benchmark stipulates that the train set and test set should include the adversarial examples come from attacking different architectures. For example, the detector is trained on the adversarial examples generated by attacking a classifier with the conv-4 network (\textit{Train Arch}), but tested on the ones of ResNet-10 (\textit{Test Arch}). Tab. \ref{tab:cross_arch_protocol} shows the detail of this benchmark, all types of attacks are used to train the detectors. Three architectures are used, namely conv-4, ResNet-10 and ResNet-18. Note that the concept of architecture in this benchmark is only related to the classifier's backbone during adversarial examples generation, but not related to the detector model.

\subsection{White-box Attack Benchmark}
\label{sec:white_box_attack_benchmark}
The \textit{white-box attack} means the adversary has the information of both the image classifier and is aware of the detector. It has full knowledge of the detector. In other words, the adversary needs to fool both the classifier and detector simultaneously, making it more challenge to defend. We use the targeted iterative FGSM (I-FGSM) \cite{kurakin2016adversarial} and C\&W \cite{Carlini2017TowardsET} attacks to simulate white-box attacks with the method presented in Carlini and Wagner \cite{Carlini2017TowardsET}. The basic idea is to construct a combined model which combines the original classifier model and the detector. The original classifier has $N$ output labels, then the new model outputs $N+1$ labels with the last label indicates whether the input is an adversarial example. More specifically, let's denote the new model as $B$ which combines the classifier $C$ and the detector $D$. $B$'s output logits is denoted as $Z_B$, $C$'s output is denoted as $Z_C$ and $D$'s output as $Z_D$. The $Z_B$ is constructed using the following formula:
\begin{equation}
\label{eqn:white_box_attack}
Z_B(x)_i = \begin{cases}
Z_C(x)_i & \text{if }i \le N \\
Z_D(x) \times 2 \times \text{max} Z_C(x) & \text{if }i=N+1
\end{cases}
\end{equation}
It is easy to see that when an input is detected as an adversarial example by $D$, then $Z_D$ would be larger than 0.5 and it leads $Z_B(x)_{N+1}$ to be larger than $Z_B(x)_i$ for $1\le i \le N$. If an input is detected as a real example, $B$ classifies it the same label as $C$ does. In this way, the new model $B$ combines $C$ and $D$.

Now, we can use the targeted iterative FGSM (I-FGSM) or C\&W adversary to attack this new model $B$ to generate the white-box adversarial example. The target label is set to make $C$ classify this example incorrectly but make the example bypass the detector $D$. In MetaAdvDet, $D$ represents for the learned master network $\mathcal{M}$ which would be attacked. Although the white-box attack leads $\mathcal{M}$ to misclassify, MetaAdvDet can benefit from the learning-to-learn strategy for recovering to the correct prediction with limited white-box examples provided, as steps shown in Algorithm \ref{alg:test}.

\section{Experiment}
\subsection{Experiment Setting}
\vspace{-0.1cm}
\begin{table}[!htb]
	\scriptsize
	\tabcolsep=0.1cm
	\setlength{\abovecaptionskip}{0pt}%
	\setlength{\belowcaptionskip}{0pt}%
	\caption{The modules of one block in the conv-3 backbone of MetaAdvDet and other compared methods. The conv-3 backbone consists of 3 such blocks in total, and the last block connects to a fully-connected layer to output a vector with two probabilities.}
	\begin{center}
		\begin{tabular}{ccc}
			\toprule
			index &	module & parameter configuration \\
			\midrule
			1 & conv-layer & 3$\times$3 kernel, channel = 64, pad = 0 \\
			2 & batch normalization & momentum = 1 \\ 
			3 & ReLU & - \\
			4 & max pooling & 2$\times$2 pooling, stride = 2 \\
			\bottomrule
		\end{tabular}
	\end{center}
	
	\label{tab:conv3}
	\vspace{-0.3cm}
\end{table}

\label{sec:meta_setting}
\begin{table}[!htb]
	\scriptsize
	\tabcolsep=0.1cm
	\setlength{\abovecaptionskip}{0pt}%
	\setlength{\belowcaptionskip}{0pt}%
	\caption{The default parameters configuration which is used in the ablation study in Sec. \ref{sec:ablation_study}, and also used in other comparative experiments of Sec. \ref{sec:expr_cross_adv}, Sec. \ref{sec:expr_cross_domain}, Sec. \ref{sec:expr_cross_arch} and Sec. \ref{sec:expr_white_box}.}
	\begin{center}
		\begin{tabular}{p{1.8cm}|p{1.4cm}|p{4.4cm}}
			\toprule
			name & default value & description \\
			\midrule
			shots &  1 & number of examples in a way, MetaAdvDet should set the same shots in both training and testing. \\
			ways & 2 & alias of class number, data of the same way come from using the same adversary to attack the same category's images. \\
			train query set size & 70 & number of examples of a query set in training. \\ 
			test query set size & 30 & number of examples of a query set in testing. \\ 
			task number $K$ & 30 & number of tasks in each mini-batch. \\
			inner update times & 12 & iteration times of inner update during training  \\
			fine-tune times & 20 & iteration times of fine-tune during testing. \\
			total tasks & 20000 & total tasks in the constructed tasks. \\
			inner learning rate & 0.001 & learning rate of inner update. \\
			outer learning rate & 0.0001 & learning rate of outer update. \\
			dataset & AdvCIFAR & the dataset for ablation study \\
			backbone & conv-3 & the backbone of MetaAdvDet \& compared methods \\
			benchmark & cross-adversary& the benchmark for ablation study \\
			\bottomrule
		\end{tabular}
	\end{center}
	\label{tab:default}
	\vspace{-0.1cm}
\end{table}

In the construction of tasks, we set the task number to be \nn{20000} in total, which covers all the samples of the original datasets.
During the learning process, 30 tasks are randomly chosen from these tasks to form each mini-batch. 
In each task, the two-way setting is applied which makes MetaAdvDet to be a binary classifier for distinguishing the adversarial examples from real examples, as shown in Fig. \ref{fig:arch}.  The inner-update learning rate $\lambda_1$ is set to $0.001$ empirically. Because of the summation of gradients in Algorithm \ref{alg:trn}, the outer-update learning rate $\lambda_2$ is set to $0.0001$ which is 10 times smaller than $\lambda_1$. The training epoch is set to 4, because after 4 epochs, we observe that the F1 score on the validation set is stable. The query set size used for outer-update is set to 70 for two ways, that is 35 samples in each way. The fine-tune iteration times is set to 20 which reaches the stable performance (Fig. \ref{fig:finetune_study}). 
All parameters configuration is shown in Tab. \ref{tab:default} which is set empirically based on validation set.

\subsection{Evaluation Metric}
\label{sec:eval_metric}
Our metric restricts that all compared methods need to be evaluated on 1000 testing tasks to cover all test samples. To quantify the detection performance of all detection methods, we adopt the F1 score which follows Liang \etal \cite{liang2018detecting} and Sabokrou \etal \cite{sabokrou2018adversarially}. It is defined as the harmonic mean between \textit{precision} and \textit{recall}:

\begin{equation}
\label{eqn:F1}
\begin{aligned}
&\text{recall} = \frac{TP}{TP+FN}, \text{precision} = \frac{TP}{TP+FP} \\
&\text{F1} = 2 \times \frac{\text{precision} \times \text{recall}}{\text{precision} + \text{recall}}
\end{aligned}
\end{equation}
We use label 1 to represent the real example and 0 to represent the adversarial example, so \textit{TP} is the number of correctly detected real examples, \textit{FN} is the number of real examples that are incorrectly detected as adversarial examples, and \textit{FP} is the number of adversarial images that are detected as real examples. Note that the final F1 score is obtained via averaging F1 scores of all tasks (Algorithm \ref{alg:test}).

\subsection{Compared Methods}

The selection of compared state-of-the-art methods are based on the consideration of two principles: (1) In order to comply with the few-shot-fashion benchmarks, the compared method must be an end-to-end learning approach to be fine-tuned in test stage. 
(2) The compared methods are able to detect new attacks, in order to evaluate and compare the detection technique in terms of evolving adversarial attacks. Based on above principles, MetaAdvDet is compared with a image rotation transformation based detector, named \textit{TransformDet} \cite{tian2018detecting}; and a detection technique based on a secret fingerprint, named \textit{NeuralFP} \cite{dathathri2018detecting}. The NeuralFP is trained for 100 epochs on each dataset, and TransformDet is trained for 10 epochs on each dataset. In the fine-tune step, because NeuralFP is trained on real examples, we extract the real examples from support set to fine-tune. In addition, NeuralFP obtains the F1 score via determining the best threshold for each task. We configure these methods following their original settings \cite{tian2018detecting,dathathri2018detecting} (Tab. \ref{tab:protocol_conf}).

\begin{table}[htp]
	\scriptsize	
	\centering
	\tabcolsep=0.15cm
	\caption{The configuration of train, validation and test set of all compared methods on the proposed benchmarks.}
	\begin{tabular}{p{1.6cm}|p{2.1cm}|p{1.7cm} | p{1.7cm}}
		\toprule
		Method & Train Set & Test Set & Validation Set\\
		\hline
		DNN & train set of adversarial example datasets (\textit{e.g.} AdvCIFAR \textit{etc.})  & \multirow{5}{1.7cm}{constructed tasks of test set in adversarial example datasets, each task contains the support set and the query set. The performance is evaluated on the query set.} &  \multirow{5}{1.7cm}{constructed tasks of validation set in adversarial example datasets, each task contains the support set and the query set. The performance is evaluated on the query set.} \\
		\cline{1-2}
		DNN (balanced) & train set of adversarial example datasets which is down-sampling to keep class balanced & & \\
		\cline{1-2}
		NeuralFP \cite{dathathri2018detecting} & real examples of train set in the original dataset (\textit{e.g.} CIFAR-10 \textit{etc.}) & & \\
		\cline{1-2}
		TransformDet \cite{tian2018detecting} & train set of adversarial example datasets, down-sampling if necessary. & & \\
		\cline{1-2}
		MetaAdvDet & constructed tasks of train set in adversarial example datasets & & \\
		\bottomrule
	\end{tabular}
	\label{tab:protocol_conf}
\end{table}

We adopt a binary neural network classifier as the baseline, denotes as \textit{DNN}. DNN is trained on all data of adversarial example dataset and its backbone is the same with MetaAdvDet, which is a 3 conv-layers network (Tab. \ref{tab:conv3}). Because the adversarial example datasets are the highly class imbalance datasets which contain much more adversarial examples than real ones, we train the other DNN by using the balanced data between adversarial and real samples by down-sampling, which is denoted as \textit{DNN (balanced)}. The dataset configurations of different methods are listed in Tab. \ref{tab:protocol_conf}.

\subsection{Ablation Study}
\label{sec:ablation_study}

\begin{figure}[htbp]
	\setlength{\abovecaptionskip}{0pt}
	\setlength{\belowcaptionskip}{0pt}
	\captionsetup[sub]{font={scriptsize}}
	\centering 
	\begin{minipage}[b]{.23\textwidth} 
		
		\includegraphics[width=\linewidth]{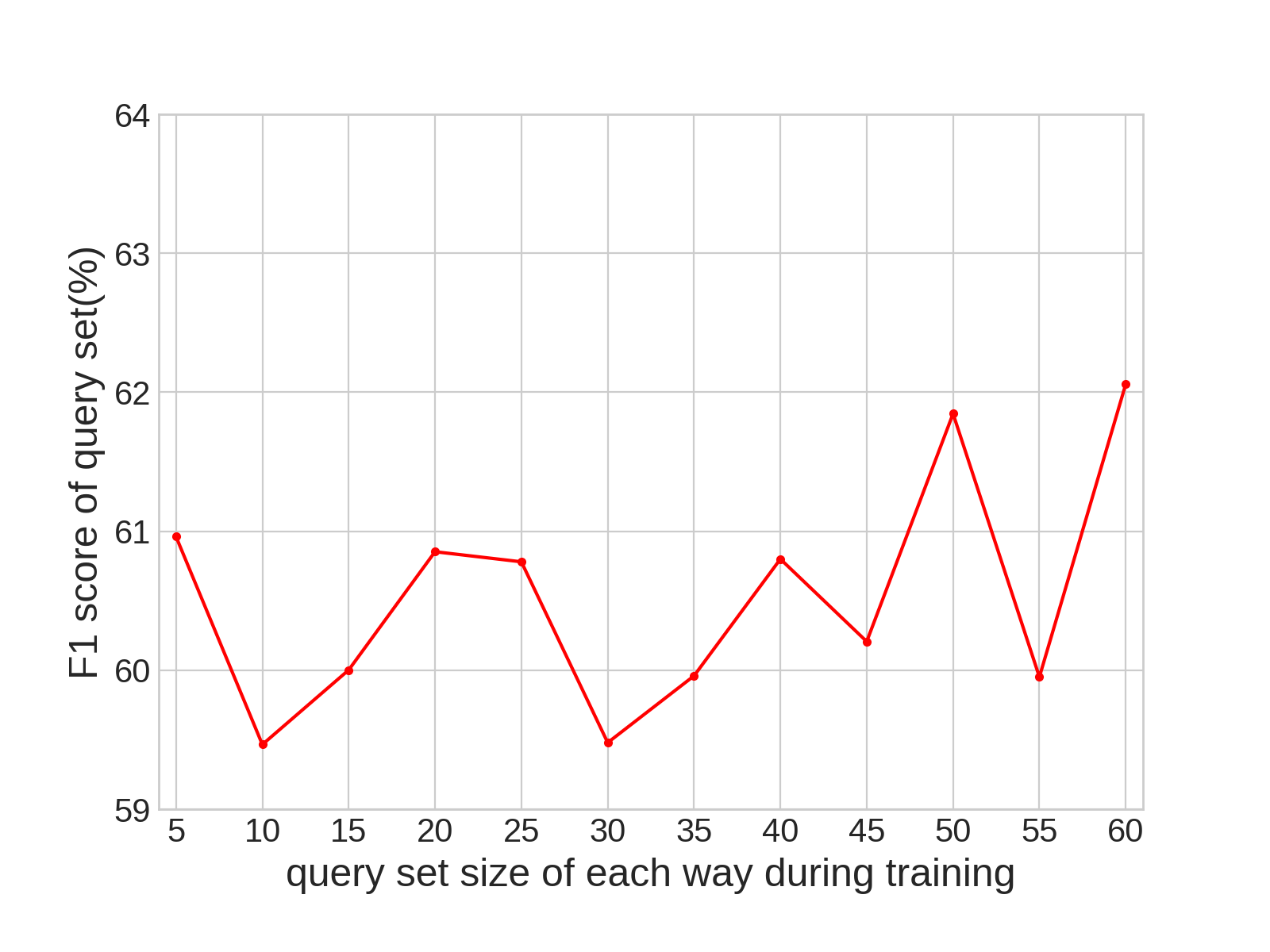}
		\subcaption{train query set size study}
		\label{fig:query_size}
	\end{minipage}
	\begin{minipage}[b]{.23\textwidth}
		\includegraphics[width=\linewidth]{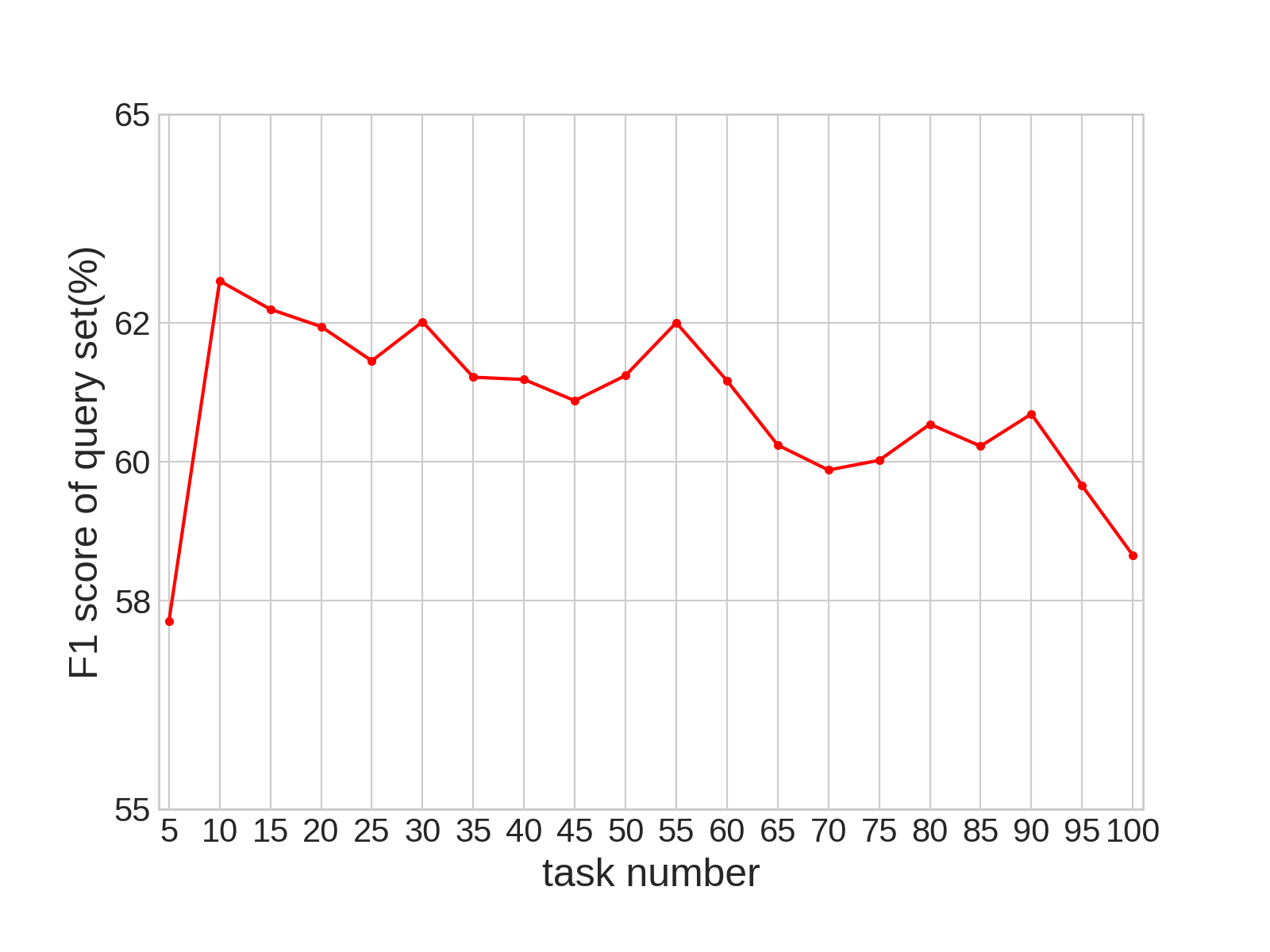}
		\subcaption{task number $K$ study}
		\label{fig:task_number}
	\end{minipage}
	\caption{Ablation study results of train query set size and task number of a training mini-batch.}
	\label{fig:ablation_1}
	\vspace{-0.5cm}
\end{figure}

\begin{figure}[htbp]
	\setlength{\abovecaptionskip}{0pt}
	\setlength{\belowcaptionskip}{0pt}
	\captionsetup[sub]{font={scriptsize}}
	\centering 
	\begin{minipage}[b]{.23\textwidth}
		\includegraphics[width=\linewidth]{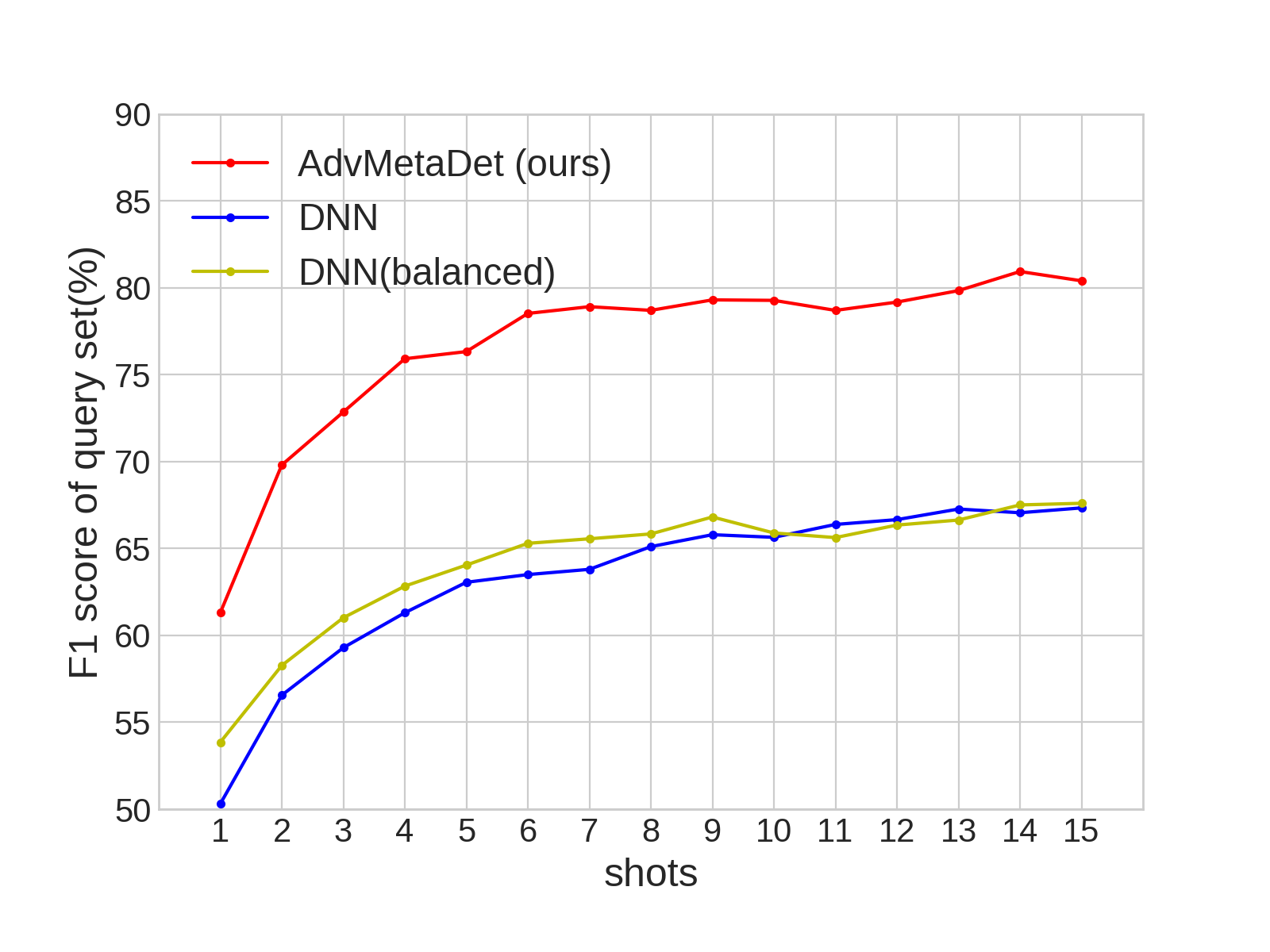}
		\subcaption{shots study}
		\label{fig:shots_study}
	\end{minipage}
	\begin{minipage}[b]{.23\textwidth}
		\includegraphics[width=\linewidth]{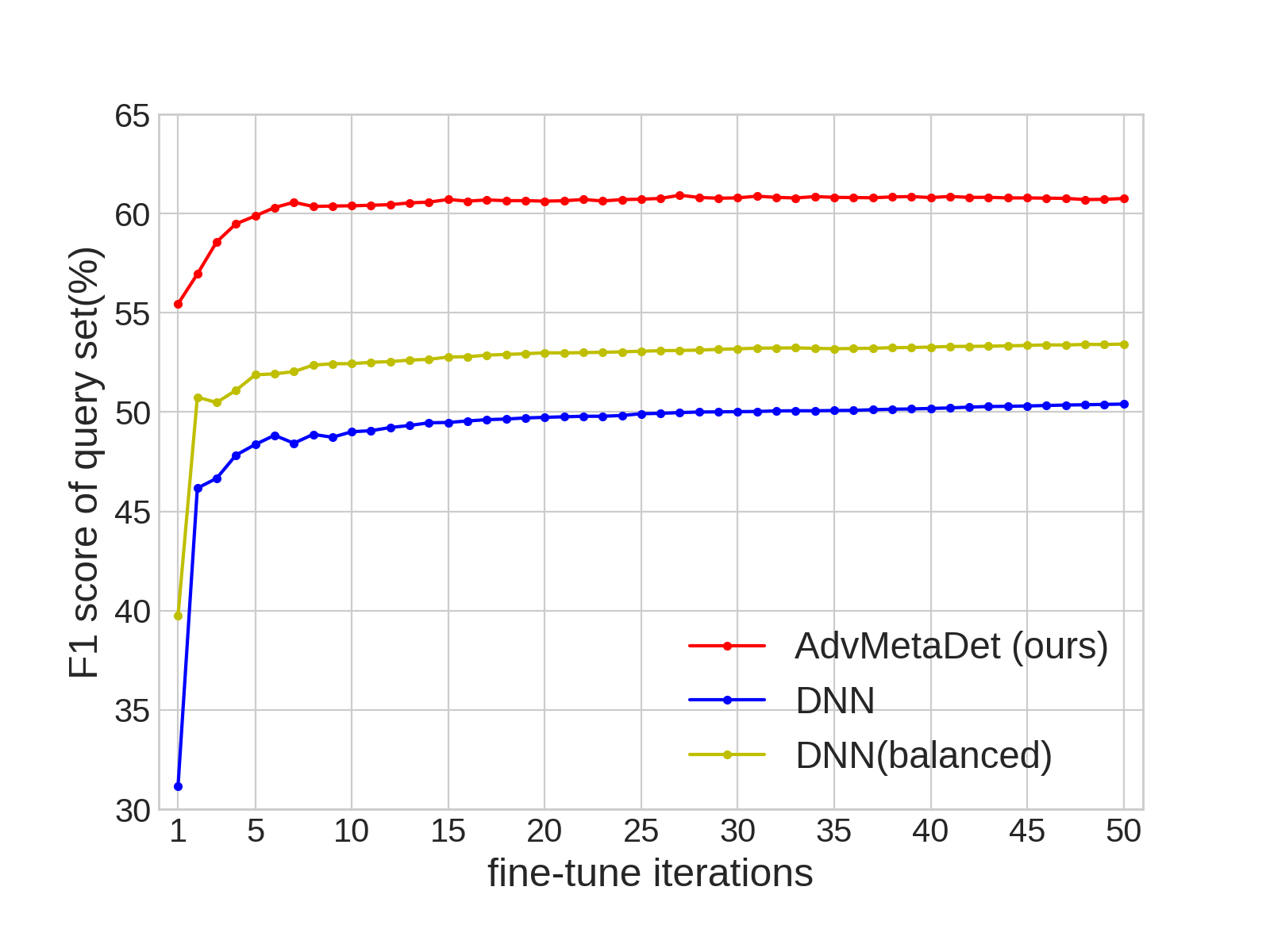}
		\subcaption{fine-tune iterations study}
		\label{fig:finetune_study}
	\end{minipage}
	\caption{Ablation study results of shots and fine-tune iterations. MetaAdvDet outperforms the baseline DNN and DNN (balanced) by a large margin.}
	\label{fig:ablation_2}
	\vspace{-0.3cm}
\end{figure}

To inspect the effect of each key parameter respectively, we conduct the control experiments on AdvCIFAR by adjusting one parameter while keeping other parameters fixed as listed Tab. \ref{tab:default}.  Fig. \ref{fig:ablation_1} and Fig. \ref{fig:ablation_2} are the results of the cross-adversary benchmark.

\newrobustcmd{\B}{\bfseries}
\vspace{-0.4cm}

From Fig. \ref{fig:ablation_2}, following conclusions can be drawn:

1) MetaAdvDet outperforms DNN with only a few fine-tuning iterations, \textit{e.g.} MetaAdvDet even surpasses the results of all fine-tune iterations of DNN by only using a single iteration (Fig. \ref{fig:finetune_study}).

2) The balanced training data of DNN (balanced) helps to improve performance over DNN with the few-shot fine-tunings (Fig. \ref{fig:finetune_study}).

\begin{table}[htb]
	\scriptsize
	\tabcolsep=0.1cm
	\setlength{\abovecaptionskip}{0pt}%
	\setlength{\belowcaptionskip}{0pt}%
	\caption{F1 score of randomized-way and fixed-way settings in AdvCIFAR. The randomized-way indicates that the labels of two ways are shuffled in each task. The fixed-way uses label 1 as real example and label 0 as adversarial example.}
	\begin{center}
		\begin{tabular}{c|c|c}
			\toprule
			shots & fixed-way & randomized-way \\
			\midrule
			1 & \B 0.686 & 0.616 \\
			5 & \B 0.787  & 0.772 \\ 
			\bottomrule
		\end{tabular}
	\end{center}
	\label{tab:random_vs_fix_way}
	\vspace{-0.3cm}
\end{table}
Tab. \ref{tab:random_vs_fix_way} illustrates the F1 score results of a randomized-way and fixed-way assignment settings in AdvCIFAR dataset. It shows that the result of fixed-way setting outperforms that of randomized-way setting. In following experiments, we use the fixed-way setting.

\subsection{Cross-Adversary Benchmark Result}
\label{sec:expr_cross_adv}
To compare the performance of our approach and other state-of-the-art methods under the cross-adversary benchmark. In this section, we collect the results of TransformDet \cite{tian2018detecting}, NeuralFP \cite{dathathri2018detecting}, baseline DNN and DNN (balanced). Tab. \ref{tab:cross_adversary_result} shows that MetaAdvDet outperforms the baseline and other methods in nearly all datasets. Thus, we can conclude MetaAdvDet is able to achieve high performance in detecting new attack with limited examples of that attack. 

Our approach is particularly effective in detecting the attacks that exhibit quite different appearance from the training attacks. Typical representative attacks are Spatial Transformation, \textit{etc.}, the results of three representative attacks are shown in Tab. \ref{tab:stats_for_each_attack_cross_adversary_CIFAR10}.

\begin{table}[!htb]
	\scriptsize
	\tabcolsep=0.1cm
	\setlength{\abovecaptionskip}{0pt}%
	\setlength{\belowcaptionskip}{0pt}%
	\caption{F1 score of the cross-adversary benchmark, this table shows the results in using the adversarial examples generated by attacking the classifier with conv-4 architecture.}
	\begin{center}
		\begin{tabular}{c|p{2.4cm}|cc}
			\toprule
			\multirow{2}*{Dataset} & \multicolumn{1}{c|}{\multirow{2}*{Method}} & \multicolumn{2}{c}{F1 score} \\
			\cmidrule(rl){3-4} & &1-shot & 5-shot \\  
			\midrule
			\multirow{5}{*}{AdvCIFAR} 
			& DNN & 0.495 & 0.639 \\
			& DNN (balanced) & 0.536 & 0.643 \\
			& NeuralFP \cite{dathathri2018detecting} & \B 0.698 & 0.700 \\
			& TransformDet \cite{tian2018detecting} & 0.662 & 0.697 \\
			& MetaAdvDet (ours) & 0.685 & \B 0.791 \\
			\midrule
			\multirow{5}{*}{AdvMNIST} 
			& DNN & 0.812 & 0.852 \\
			& DNN (balanced) & 0.797  & 0.808 \\
			& NeuralFP \cite{dathathri2018detecting} & 0.780  & 0.906  \\
			& TransformDet \cite{tian2018detecting} & 0.840  & 0.904 \\
			& MetaAdvDet (ours) & \B 0.987  & \B 0.993 \\
			\midrule
			\multirow{5}{*}{AdvFashionMNIST} 
			& DNN & 0.782 & 0.885 \\
			& DNN (balanced) & 0.744  & 0.850 \\
			& NeuralFP \cite{dathathri2018detecting} & 0.798  & 0.817  \\
			& TransformDet \cite{tian2018detecting} & 0.712  & 0.879 \\
			& MetaAdvDet (ours) & \B 0.848  & \B 0.944 \\
			
			\bottomrule
		\end{tabular}
	\end{center}
	
	\label{tab:cross_adversary_result}
	\vspace{-0.4cm}
\end{table}

\begin{table}[!htb]
	\scriptsize
	\tabcolsep=0.1cm
	\setlength{\abovecaptionskip}{0pt}%
	\setlength{\belowcaptionskip}{0pt}%
	\caption{F1 score of representative adversaries on the AdvCIFAR dataset, cross-adversary benchmark.}
	\begin{center}
		\begin{tabular}{c|l|l|cc}
			\toprule
			\multirow{2}*{Dataset} & \multicolumn{1}{c|}{\multirow{2}*{Adversary}} & \multicolumn{1}{c|}{\multirow{2}*{Method}} & \multicolumn{2}{c}{F1 score} \\
			\cmidrule(rl){4-5} & & & 1-shot & 5-shot \\  
			\midrule
			\multirow{20}{*}{AdvCIFAR} & \multirow{5}{*}{Spatial Transformation \cite{xiao2018spatially}} & DNN & 0.498 & 0.599 \\
			& & DNN (balanced)  & 0.529 & 0.589 \\
			& & NeuralFP \cite{dathathri2018detecting} & 0.708 & 0.696 \\
			& & TransformDet \cite{tian2018detecting} & 0.633 & 0.660 \\
			& & MetaAdvDet (ours) & \B 0.811 & \B 0.920 \\
			\cmidrule(l{0.3\tabcolsep}){2-5}
			& \multirow{5}{*}{semantic \cite{hosseini2017limitation}} & DNN & 0.488 & 0.644 \\
			& & DNN (balanced)  & 0.529 & 0.657 \\
			& & NeuralFP \cite{dathathri2018detecting} & 0.698 & 0.700 \\
			& & TransformDet \cite{tian2018detecting} & 0.662 & 0.688 \\
			& & MetaAdvDet (ours) & \B 0.763 & \B 0.855 \\
			\cmidrule(l{0.3\tabcolsep}){2-5}
			& \multirow{5}{*}{NewtonFool \cite{jang2017objective}} & DNN & 0.511 & 0.664 \\
			& & DNN (balanced)  & 0.542 & 0.670 \\
			& & NeuralFP \cite{dathathri2018detecting} & \B 0.696 & 0.696 \\
			& & TransformDet \cite{tian2018detecting} & 0.658 & 0.716 \\
			& & MetaAdvDet (ours) & 0.647 & \B 0.735 \\
			\bottomrule
		\end{tabular}
	\end{center}
	
	\label{tab:stats_for_each_attack_cross_adversary_CIFAR10}
	\vspace{-0.4cm}
\end{table}

\subsection{Cross-Domain Benchmark Result}
\label{sec:expr_cross_domain}
\begin{table}[!htb]
	\scriptsize
	\tabcolsep=0.1cm
	\setlength{\abovecaptionskip}{0pt}%
	\setlength{\belowcaptionskip}{0pt}%
	\caption{F1 score of the cross-domain benchmark, this table shows the results which are evaluated on the adversarial examples generated by attacking the conv-4 network. All the types of attacks are used to train the detectors.}
	\begin{center}
		\begin{tabular}{c|c|l|cc}
			\toprule
			\multirow{2}*{Train Domain} & \multirow{2}*{Test Domain} & \multicolumn{1}{c|}{\multirow{2}*{Method}} & \multicolumn{2}{c}{F1 score} \\
			\cmidrule(rl){4-5} & & & 1-shot & 5-shot \\  
			\midrule

			\multirow{4}{*}{AdvMNIST} & \multirow{4}{*}{AdvFashionMNIST} & DNN (balanced) & 0.698 & 0.813 \\
			& & NeuralFP \cite{dathathri2018detecting} & 0.748 & 0.811 \\
			& & TransformDet \cite{tian2018detecting} & 0.664 & 0.808 \\
			& & MetaAdvDet (ours) & \B 0.799 & \B 0.870 \\
			\midrule
			
			\multirow{4}{*}{AdvFashionMNIST} & \multirow{4}{*}{AdvMNIST} & DNN (balanced) & 0.950 & 0.977 \\
			& & NeuralFP \cite{dathathri2018detecting} & 0.775 & 0.836 \\
			& & TransformDet \cite{tian2018detecting} & 0.934 & 0.940 \\
			& & MetaAdvDet (ours) & \B 0.956 & \B 0.981 \\
			\bottomrule
		\end{tabular}
	\end{center}
	\label{tab:cross_domain}
	\vspace{-0.3cm}
\end{table}

In the cross-domain benchmark, the models are trained in one domain, and tested in the other domain's test set (Sec. \ref{sec:cross_domain}). 
We use DNN (balanced) instead of DNN in this benchmark because the all types of attacks are used to train which results in the highly imbalanced data classification issue if using DNN. Tab. \ref{tab:cross_domain} shows the result, which demonstrates that MetaAdvDet has an advantage in hard test set. For example, when training on AdvMNIST and testing on AdvFashionMNIST, MetaAdvDet outperforms DNN (balanced) by a large margin (\textbf{10.1\%} improvement in 1-shot).

\begin{table}[!htb]
	\scriptsize
	\tabcolsep=0.1cm
	\setlength{\abovecaptionskip}{0pt}%
	\setlength{\belowcaptionskip}{0pt}%
	\caption{F1 score of cross-architecture benchmark.}
	\begin{center}
		\begin{tabular}{c|cc|c|cc}
			\toprule
			\multirow{2}*{Dataset} & \multirow{2}*{Train Arch} & \multirow{2}*{Test Arch} & \multirow{2}*{Method} & \multicolumn{2}{c}{F1 score} \\
			\cmidrule(rl){5-6} & & & & 1-shot & 5-shot \\  
			\midrule
			
			\multirow{18}*{AdvCIFAR} & \multirow{4}*{ResNet-10} & \multirow{4}*{ResNet-18} & NeuralFP \cite{dathathri2018detecting} & 0.713 & 0.709 \\
			& & & TransformDet \cite{tian2018detecting} & 0.758 & 0.880 \\
			& & & DNN (balanced) & 0.702 & 0.768 \\
			& & & MetaAdvDet (ours) & \B 0.832 & \B 0.902 \\
			\cmidrule(r){2-6}
			& \multirow{4}*{ResNet-18} & \multirow{4}*{ResNet-10} & NeuralFP \cite{dathathri2018detecting} & 0.712 & 0.703 \\
			& & & TransformDet \cite{tian2018detecting} & 0.788 & 0.874 \\
			& & & DNN (balanced) & 0.711 & 0.752 \\
			& & & MetaAdvDet (ours) & \B 0.840 & \B 0.889 \\
			\cmidrule(r){2-6} 
			& \multirow{4}*{conv-4} & \multirow{4}*{ResNet-10} & NeuralFP \cite{dathathri2018detecting} & 0.712 & 0.703 \\
			& & & TransformDet \cite{tian2018detecting} & 0.763 & 0.868 \\
			& & & DNN (balanced) & 0.723 & 0.779 \\
			& & & MetaAdvDet (ours) & \B 0.835 & \B 0.885 \\
			\cmidrule(r){2-6}
			& \multirow{4}*{ResNet-10} & \multirow{4}*{conv-4} & NeuralFP \cite{dathathri2018detecting} & 0.709 & 0.702 \\
			& & & TransformDet \cite{tian2018detecting} & 0.766 & 0.885 \\
			& & & DNN (balanced) & 0.739 & 0.790 \\
			& & & MetaAdvDet (ours) & \B 0.854 & \B 0.918 \\
			
			\midrule
			\multirow{18}*{AdvMNIST} & \multirow{4}*{ResNet-10} & \multirow{4}*{ResNet-18} & NeuralFP \cite{dathathri2018detecting} & 0.906 & 0.882 \\
			& & & TransformDet \cite{tian2018detecting} & 0.973 & 0.988 \\
			& & & DNN (balanced) & 0.943 & 0.972 \\
			& & & MetaAdvDet (ours) & \B 0.984 & \B 0.993 \\
			\cmidrule(r){2-6}
			& \multirow{4}*{ResNet-18} & \multirow{4}*{ResNet-10} & NeuralFP \cite{dathathri2018detecting} & 0.894 & 0.738 \\
			& & & TransformDet \cite{tian2018detecting} & 0.967 & 0.990 \\
			& & & DNN (balanced) & 0.912 & 0.953 \\
			& & & MetaAdvDet (ours) & \B 0.981 & \B 0.991 \\
			\cmidrule(r){2-6}
			& \multirow{4}*{conv-4} & \multirow{4}*{ResNet-10} & NeuralFP \cite{dathathri2018detecting} & 0.894 & 0.738 \\
			& & & TransformDet \cite{tian2018detecting} & \B 0.972 & \B 0.985 \\
			& & & DNN (balanced) & 0.897 & 0.959 \\
			& & & MetaAdvDet (ours) & 0.963 & 0.983 \\
			\cmidrule(r){2-6}
			& \multirow{4}*{ResNet-10} & \multirow{4}*{conv-4} & NeuralFP \cite{dathathri2018detecting} & 0.917 & 0.961 \\
			& & & TransformDet \cite{tian2018detecting} & 0.984 & 0.992 \\
			& & & DNN (balanced) & 0.958 & 0.978 \\
			& & & MetaAdvDet (ours) & \B 0.990 & \B 0.996 \\
			
			\midrule
			\multirow{18}*{AdvFashionMNIST} & \multirow{4}*{ResNet-10} & \multirow{4}*{ResNet-18} & NeuralFP \cite{dathathri2018detecting} & 0.813 & 0.856 \\
			& & & TransformDet \cite{tian2018detecting} & 0.936 & 0.974 \\
			& & & DNN (balanced) & 0.848 & 0.932 \\
			& & & MetaAdvDet (ours) & \B 0.960 & \B 0.979 \\
			\cmidrule(r){2-6}
			& \multirow{4}*{ResNet-18} & \multirow{4}*{ResNet-10} & NeuralFP \cite{dathathri2018detecting} & 0.820 & 0.838 \\
			& & & TransformDet \cite{tian2018detecting} & 0.935 & 0.972 \\
			& & & DNN (balanced) & 0.829 & 0.918 \\
			& & & MetaAdvDet (ours) & \B 0.957 & \B 0.976 \\
			\cmidrule(r){2-6}
			& \multirow{4}*{conv-4} & \multirow{4}*{ResNet-10} & NeuralFP \cite{dathathri2018detecting} & 0.820 & 0.838 \\
			& & & TransformDet \cite{tian2018detecting} & 0.946 & 0.970 \\
			& & & DNN (balanced) & 0.920 & 0.968 \\
			& & & MetaAdvDet (ours) & \B 0.946 & \B 0.975 \\
			\cmidrule(r){2-6}
			& \multirow{4}*{ResNet-10} & \multirow{4}*{conv-4} & NeuralFP \cite{dathathri2018detecting} & 0.817 & 0.911\\
			& & & TransformDet \cite{tian2018detecting} & 0.945 & 0.979 \\
			& & & DNN (balanced) & 0.886 & 0.945 \\
			& & & MetaAdvDet (ours) & \B 0.967 & \B 0.982 \\
			
			\bottomrule
		\end{tabular}
	\end{center}
	\label{tab:cross_arch}
	\vspace{-0.5cm}
\end{table}
\subsection{Cross-Architecture Benchmark Result}
\label{sec:expr_cross_arch}

Tab. \ref{tab:cross_arch} shows the results of cross-architecture benchmark. Because NeuralFP is trained on the real samples, thus the same NeuralFP model is tested on the examples of different test architectures (Test Arch). Tab. \ref{tab:cross_arch} shows that MetaAdvDet outperforms other methods under different train and test architecture combinations, proving the superiority of MetaAdvDet in the cross-architecture benchmark.

\subsection{White-box Attack Benchmark Result}
\label{sec:expr_white_box}
In Tab. \ref{tab:white_box_result}, we present the detection performance of the white-box benchmark. The NeuralFP \cite{dathathri2018detecting} result is omitted because it detects the attack by setting threshold rather than conducting classification, which cannot be used in the method of Sec. \ref{sec:white_box_attack_benchmark}. Tab. \ref{tab:white_box_result} shows that: (1) MetaAdvDet can effectively detect white-box attack even with only one white-box example provided. (2) White-box attack targets on the master network of the meta-learner in MetaAdvDet, whereas it targets on the detector itself in other methods. 

\begin{table}[!htb]
	\scriptsize
	\tabcolsep=0.1cm
	\setlength{\abovecaptionskip}{0pt}%
	\setlength{\belowcaptionskip}{0pt}%
	\caption{F1 score of white-box attack benchmark.}
	\begin{center}
		\begin{tabular}{c|c|cc|cc}
			\toprule
			\multirow{3}*{Dataset} & \multirow{3}*{Method} & \multicolumn{2}{c|}{I-FGSM Attack} & \multicolumn{2}{c}{C\&W Attack} \\
			\cmidrule(rl){3-6}
			&  & 1-shot & 5-shot & 1-shot & 5-shot \\
			\midrule
			
			\multirow{3}*{CIFAR-10} & DNN (balanced) & 0.466 & 0.537 & 0.459 & 0.527 \\
			& TransformDet \cite{tian2018detecting} & \B 0.593 & \B 0.728 & 0.443 & 0.502 \\
			& MetaAdvDet (ours) & 0.553 &  0.633 & \B 0.548 & \B 0.607 \\		
			\midrule
			
			\multirow{3}*{MNIST} & DNN (balanced) & 0.857 & 0.956 & 0.814 & 0.913 \\
			& TransformDet \cite{tian2018detecting} & 0.864 & 0.952 & 0.775 & 0.893 \\
			& MetaAdvDet (ours) & \B 0.968 & \B 0.994 & \B 0.920 & \B 0.990 \\		
			\midrule
			
			\multirow{3}*{FashionMNIST} & DNN (balanced) & 0.745 & 0.890 & 0.726 & 0.853 \\
			& TransformDet \cite{tian2018detecting} & 0.837 & 0.920 & 0.747 & 0.853 \\
			& MetaAdvDet (ours) & \B 0.849 & \B 0.963 & \B 0.882 & \B 0.967 \\		
			\bottomrule
			
		\end{tabular}
	\end{center}
	\label{tab:white_box_result}
	\vspace{-0.5cm}
\end{table}

\subsection{Inference Time}
\vspace{-0.2cm}
\begin{table}[htb]
	\scriptsize
	\tabcolsep=0.1cm
	\setlength{\abovecaptionskip}{0pt}%
	\setlength{\belowcaptionskip}{0pt}%
	\caption{The inference time(ms) of all methods.}
	\begin{center}
		\begin{tabular}{c|cccc}
			\toprule
			Method & DNN & NeuralFP \cite{dathathri2018detecting} & TransformDet \cite{tian2018detecting} & MetaAdvDet (ours) \\
			\midrule
			Inference time (ms) & $1.53 \pm 0.01$ & $2185.12 \pm 18.10$ & $69.17 \pm 2.97$ & $4.07 \pm 4.40$ \\
			\bottomrule
		\end{tabular}
	\end{center}
	\label{tab:inference_time}
	\vspace{-0.4cm}
\end{table}

We further evaluate the inference time (excluding fine-tune steps) of all the methods measured in millisecond on one NVIDIA Geforce GTX 1080Ti GPU in Tab. \ref{tab:inference_time}. It shows that MetaAdvDet obtains the comparable inference time to DNN, due to that both methods use the same network architecture and the same feed-forward procedure for inference. In contrast, TransformDet applies multiple transformations on the input image which increases the inference time. NeuralFP tests multiple thresholds to determine the best threshold for detection, which significantly increases the inference time.

\section{Conclusion}
In this paper, we present a meta-learning based adversarial attack detection approach for detecting evolving adversary attacks with limited examples. To this end, the approach is equipped with a double-network framework which includes a task-dedicated network and a master network to learn from either individual tasks or the task distribution. In this way, the rapid adaption capability of detecting new attacks is achieved. Experimental results conclude that:
(1) Tab. \ref{tab:cross_adversary_result}, Tab. \ref{tab:cross_domain} and Tab. \ref{tab:cross_arch} show that NeuralFP gets lower F1 scores than ours under different benchmarks. It manifests that NeuralFP which is trained on real examples cannot detect evolving attacks effectively.
(2) We get the lowest results in AdvCIFAR dataset (Tab. \ref{tab:cross_adversary_result}, Tab. \ref{tab:cross_arch} and Tab. \ref{tab:white_box_result}), which manifests the adversarial examples generated in CIFAR-10 are more difficult to detect.
(3) MetaAdvDet performs well in the benchmarks of cross-adversary (Tab. \ref{tab:cross_adversary_result}), cross-domain (Tab. \ref{tab:cross_domain}), cross-architecture (Tab. \ref{tab:cross_arch}) and white-box attack (Tab. \ref{tab:white_box_result}), proving that MetaAdvDet is a suitable method for detecting evolving attacks with limited examples.

\clearpage
\balance 
\bibliographystyle{ACM-Reference-Format}
\bibliography{reference}

\end{document}